\documentclass[conference]{IEEEtran}
\usepackage{times}

\usepackage[numbers]{natbib}
\usepackage[bookmarks=true]{hyperref}
\usepackage{url}
\usepackage{tikz} 
\usepackage{amsmath}
\usepackage{amssymb}
\usepackage{multirow}
\usepackage{multicol}
\usepackage{mathrsfs}
\usepackage{etoolbox}
\usepackage{xspace}
\usepackage{xcolor}
\usepackage{graphicx}
\usepackage{caption}
\usepackage{subcaption}
\usepackage{adjustbox}
\usepackage{xfrac}
\usepackage{listings}
\usepackage{float}
\usepackage{booktabs}
\usepackage{bm}
\usepackage{amsfonts}

\usepackage{textcomp}
\usepackage{makecell}

\definecolor{deepblue}{HTML}{047bff}

\newcommand{\ourwebsite}[0]{\href{https://immvlab.github.io/DVS/}{https://immvlab.github.io/DVS/}}

\begin{document}

\title{Demonstrating DVS: Dynamic Virtual-Real Simulation Platform for Mobile Robotic Tasks}
\vspace{-0.2in}

\author{Zijie Zheng$^{1*}$\quad Zeshun Li$^{1*}$\quad Yunpeng Wang$^{1*}$\quad Qinghongbing Xie$^{1}$\quad Long Zeng$^{1\dagger}$\vspace{0.03in}\\

$^1$Tsinghua University\quad $^*$Equal contribution\quad$^\dagger$Corresponding author\vspace{0.1in}\\

\href{https://immvlab.github.io/DVS/}{\color{deepblue}\textbf{Dynamic Virtual Real Simulation Platform Website}\xspace}\vspace{-0.1in}}

\IEEEpeerreviewmaketitle
\twocolumn[{%
    \renewcommand\twocolumn[1][]{#1}%
        \maketitle
        \vspace{-5mm}
	\begin{center}

\vspace{-0cm}
    \includegraphics[width=18cm]{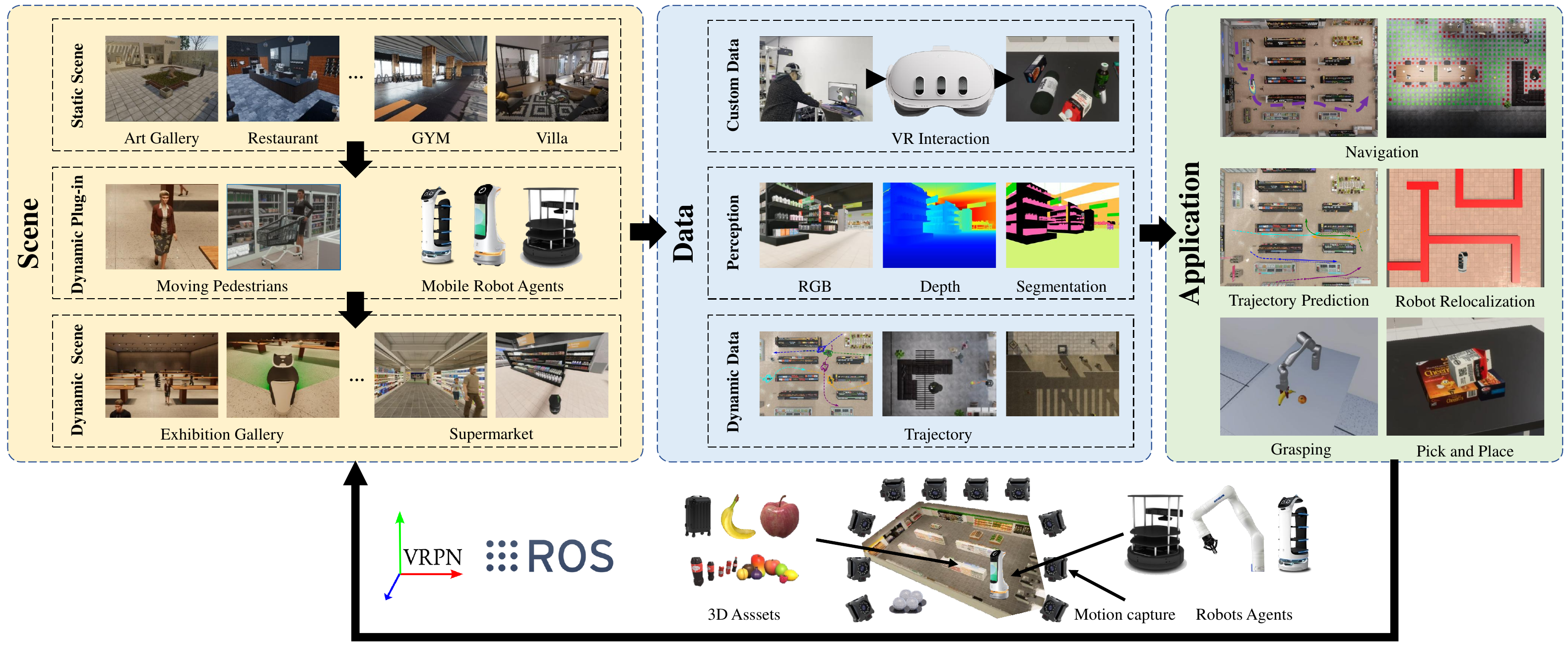}
    \captionof{figure}{Overview of DVS platform, which offers a variety of large-scale indoor scene types and dynamic element plugins on the left, enabling users to construct dynamic environments. In the middle, the platform supports various data types that can be generated, such as RGB, depth, and semantic labels. On the right, the data created using this platform can be applied to train robots for tasks such as navigation, trajectory prediction, and grasping. Through a virtual-real fusion feedback mechanism, the platform allows bidirectional mapping of the states of real and virtual agents, enriching the research scenarios.}
    \label{fig0}
    \end{center}
}]

\begin{abstract}
With the development of embodied artificial intelligence, robotic research has increasingly focused on complex tasks. Existing simulation platforms, however, are often limited to idealized environments, simple task scenarios and lack data interoperability. This restricts task decomposition and multi-task learning. Additionally, current simulation platforms face challenges in dynamic pedestrian modeling, scene editability, and synchronization between virtual and real assets. These limitations hinder real world robot deployment and feedback. To address these challenges, we propose DVS (Dynamic Virtual-Real Simulation Platform), a platform for dynamic virtual-real synchronization in mobile robotic tasks. DVS integrates a random pedestrian behavior modeling plugin and large-scale, customizable indoor scenes for generating annotated training datasets. It features an optical motion capture system, synchronizing object poses and coordinates between virtual and real world to support dynamic task benchmarking. Experimental validation shows that DVS supports tasks such as pedestrian trajectory prediction, robot path planning, and robotic arm  grasping, with potential for both simulation and real world deployment. In this way, DVS represents more than just a versatile robotic platform; it paves the way for research in human intervention in robot execution tasks and real-time feedback algorithms in virtual-real fusion environments. More information about the simulation platform is available on \ourwebsite.  
\end{abstract}

\section{Introduction}
Robot's capabilities are becoming increasingly powerful with advances in perception, decision-making, and execution technologies. These improvements have expanded their potential applications in industrial manufacturing\cite{liu2022robot,XU24JMS}, smart homes\cite{luperto2023integrating}, and other fields\cite{ParametricNet19ICRA, Language23RCIM, PathPlanning18RCIM}. The transition from rule-based operations to end-to-end learning has enabled robots to tackle more complex tasks. However, achieving high efficiency in real world scenarios requires a complete workflow: virtual data collection, simulator training, and real robot deployment. Existing simulation platforms often fail to effectively support this closed-loop research due to their functional limitations.

Data collection in robotics typically relies on two approaches: collecting real world data with physical robots or using virtual agents in simulated environments. Systems such as Mobile ALOHA\cite{fu2024mobile} aim to reduce the cost of collecting data in the real world. However, they still require significant hardware investment and expert labor. Simulators provide task-specific modeling tools for focused research. In contrast, simulation platforms offer a broader framework for multitask and complex scenario investigations, allowing faster iteration. Platforms such as Habitat\cite{szot2021habitat}, iGibson\cite{li2021igibson,shen2021igibson}, and Arena\cite{kastner2024arena} have facilitated data collection and algorithm training. Nonetheless, their applicability is often constrained by domain-specific assumptions or limited extensibility. Habitat, while extended to support HITL (Human-in-the-Loop)\cite{mosqueira2023human} and HRC (Human-Robot Collaboration)\cite{matheson2019human}, focuses mainly on navigation tasks. iGibson enhances data richness and realism through interactive environments but lacks support for dynamic scenarios. Arena specializes in navigation, while tasks like grasping rely on other simulators.

Existing simulation platforms are typically tailored to specific tasks or static environments, limiting their ability to support complex, long-horizon scenarios that demand deep environmental understanding and cross-domain collaboration. For instance, completing a task like \textit{retrieving a bottle from the fridge and placing it on a desk} involves navigation, manipulation, and environment interaction. Current methods decompose such tasks into sub-tasks across multiple simulators, increasing workload and reducing coherence. Furthermore, most platforms lack support for dynamic scenarios, such as modeling pedestrian behaviors, and do not incorporate real world feedback. This limitation exacerbates the sim-to-real gap, often leading to significant performance degradation during deployment.

We propose DVS (Dynamic Virtual-Real Simulation Platform), a novel framework tailored for dynamic, closed-loop robotic research across diverse tasks. DVS addresses existing limitations through three key features. First, it supports complex long-horizon tasks with dynamic pedestrian modeling and flexible indoor scene editing. This enables high-fidelity simulation environments for multi-stage operations. Second, it establishes a virtual-real fusion workflow, combining high-accuracy optical motion capture and ROS-based communication. This ensures synchronized validation between virtual and physical robots, facilitating optimization based on simulated feedback. Third, it introduces an intervention-based process. Researchers can adjust virtual scenarios in real-time during physical execution, enhancing task flexibility and robustness, and extending HRC research capabilities.

Key contributions of this work are as follows:
\begin{itemize}
    \item We present a virtual-real fusion simulation platform (DVS) for robotic research, which enables closed-loop sim-to-real transfer validation through virtual-physical synchronization and ROS-based communication. It supports a wide range of tasks.
    \item We provide dynamic environmental modeling, including pedestrian behavior simulation and flexible scene editing. These capabilities enhance complex task execution through diverse and high-quality data generation.
    \item We introduce an intervention-enabled workflow. This supports real-time scenario adjustments during physical deployment. The virtual-real synchronization mechanism improves adaptability in dynamic environments, demonstrated through manipulation tasks.
\end{itemize}

\begin{table*}[h]
\centering
\renewcommand{\arraystretch}{1.2} 
\setlength{\tabcolsep}{7pt} 
\caption{Comparison of Simulation Platforms. For the sensor, S refers to semantic, L refers to Lidar}
\begin{tabular}{@{}l>{\centering\arraybackslash}p{2cm}>{\centering\arraybackslash}p{2cm}cc>{\centering\arraybackslash}p{2cm}>{\centering\arraybackslash}p{2cm}@{}}
\toprule
\multirow{2}{*}{\makecell{Simulation Platform}} & \multirow{2}{*}{Sensors} & \multicolumn{2}{c}{\makecell{Dynamic Scenes}} & \multirow{2}{*}{VR Interaction} & \multirow{2}{*}{ROS} \\ \cmidrule(lr){3-4}
 &  & Pedestrians & Objects &  &  \\
\midrule
Arena\cite{kastner2024arena} & RGB-D, L & $\checkmark$ & $\times$ & $\times$ & $\checkmark$ \\
AI2THOR\cite{kolve2017ai2} & RGB-D, S & $\times$ & $\times$ & $\times$ & $\times$ \\
Gibson series\cite{li2021igibson}\cite{shen2021igibson} & RGB-D, S, L & $\times$ & $\times$ & $\checkmark$ & $\checkmark$ \\
HoME\cite{brodeur2017home} & RGB-D, S & $\times$ & $\times$ & $\times$ & $\times$ \\
Habitat\cite{savva2019habitat}\cite{szot2021habitat}\cite{puig2023habitat} & RGB-D, S & $\times$ & $\times$ & $\checkmark$ & $\times$ \\
SAPIEN\cite{xiang2020sapien} & RGB-D, S & $\times$ & $\times$ & $\checkmark$ & $\times$ \\
ThreeDWorld\cite{gan2020threedworld} & RGB-D, S & $\times$ & $\times$ & $\checkmark$ & $\times$ \\
VirtualHome\cite{puig2018virtualhome} & RGB-D, S & $\times$ & $\times$ & $\times$ & $\times$ \\
\textbf{DVS(Ours)} & \textbf{RGB-D, S, L} & \textbf{$\checkmark$} & \textbf{$\checkmark$} & \textbf{$\checkmark$} & \textbf{$\checkmark$} \\
\bottomrule
\end{tabular}
\label{tab:platform-comparison}
\end{table*}

\section{Related Work}

Simulation platforms have become integral to the development and validation of embodied artificial intelligence algorithms, enabling researchers to train and test robotic systems in controlled environments before deployment in real world tasks. These platforms have seen significant advancements over the past decade, particularly in the areas of physical modeling, scene realism, and task-specific benchmarks.

The emergence of embodied intelligence has catalyzed significant advances in robotics and AI, especially in tasks involving real world interaction and navigation. Such tasks ranging from obstacle avoidance and path planning to human-robot collaboration demand rigorous testing and training frameworks. In this context, simulation environments have emerged as indispensable tools, that offer safe, scalable, and cost-effective platforms for developing and validating embodied AI algorithms. These environments not only enable exploration of high-risk scenarios and faster algorithm iteration but also address the critical challenge of sim2real generalization, where models trained in simulation must effectively transfer to real world robotic systems.

During the past decade, the development of simulation platforms has been instrumental in advancing embodied intelligence. Platforms such as Gazebo\cite{koenig2004design}, MuJoCo\cite{todorov2012mujoco}, and NVIDIA Isaac Sim\cite{mittal2023orbit} have excelled in robotics control and high-precision physical simulation, enabling accurate modeling of robot dynamics and multi agents systems. Meanwhile, tools like Habitat\cite{puig2023habitat}, AI2-THOR\cite{kolve2017ai2}, and iGibson\cite{li2021igibson} have prioritized photorealistic environments for navigation and task planning, supporting benchmarks for tasks like object rearrangement, manipulation, and visual question answering. Recent systems such as DialFRED\cite{gao2022dialfred} and TEACh\cite{teach} have expanded the scope of these benchmarks by integrating natural language dialogue, encouraging richer agent-environment interactions. Despite these advancements, several persistent challenges remain unresolved, hindering the broader applicability of these platforms to dynamic and real world scenarios.

A major limitation of existing platforms is their inability to model dynamic, stochastic environments that capture realistic human behaviors and evolving scene conditions. Platforms like Habitat and AI2-THOR, while robust for static or semi-static environments, rely heavily on predefined tracks and scripted object interactions, which constrain their generalizability to real world, unpredictable conditions. Another challenge is the gap in the sim2real generalization. Although simulators like Pybullet\cite{coumans2016pybullet} and MuJoCo excel in physical modeling, they often lack the diversity and randomness required to robustly train algorithms for real world deployment. Moreover, the growing emphasis on human-robot collaboration has exposed the limitations of existing platforms, which rarely support real-time interactions such as gesture-based commands, shared workspaces, or natural language dialogue. Systems such as HumanTHOR\cite{wang2024demonstrating} and SEAN\cite{9851501} have made notable progress toward dynamic interaction modeling, but their focus remains limited to basic social navigation or static collaborative tasks, leaving ample room for advancement. Moreover, most existing platforms specialize either in physical modeling or in photorealistic simulation, yet rarely integrate both capabilities within a unified framework—highlighting a critical gap in tools capable of holistically supporting embodied intelligence research.

To address these limitations, we propose a novel virtual-physical integration platform that combines the strengths of high-fidelity physics, dynamic scene modeling, and real-time human-robot collaboration. By introducing stochastic pedestrian behavior modeling—including adjustable avoidance radii, randomized spawning points, and variable motion patterns—our platform supports dynamic and unpredictable environments, enhancing the robustness and generalization of robot algorithms. Additionally, a optical motion capture system provides submillimeter precision data for sim2real transfer, ensuring better deployment of simulation-trained models to real world systems. Real time human-in-the-loop (HITL) interactions\cite{mosqueira2023human}, including gesture commands, natural language dialogue, and shared workspace collaboration, further enable realistic HRC experiments. Finally, the integration of annotated synthetic data with real world motion capture enables simultaneous development and validation across both virtual and physical domains, effectively bridging the gap between simulation and reliable real world deployment in embodied AI.

By tackling key challenges in dynamic scene modeling, sim-to-real transfer, and human-robot collaboration, our proposed platform provides a unified solution for advancing embodied intelligence. Its capacity to simulate complex, real world environments and enable seamless robot deployment positions it as a powerful enabler for future research in navigation, manipulation, and collaborative tasks. A detailed comparison of existing simulation platforms is delineated in Table \ref{tab:platform-comparison}.

\section{System Framework}

\begin{figure}[h]
    \centering
    \includegraphics[width=\linewidth]{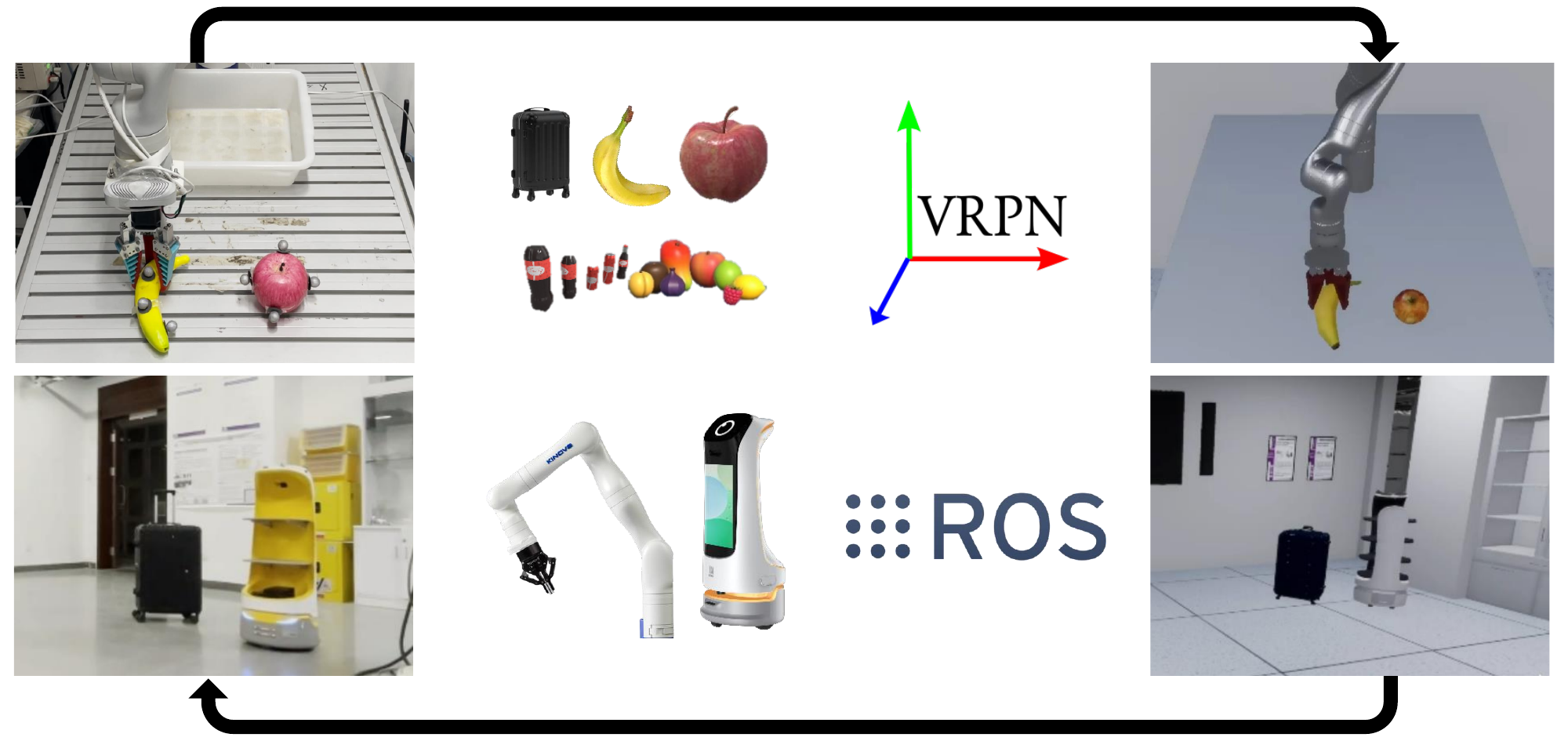}
    \caption{Virtual-Real Data Synchronization Framework. The central demonstrates the synchronization of object pose and robot motion through VRPN and ROS. The left and right parts depict the virtual simulation environment and physical real world scene, respectively.}
    \label{fig_virtual-real-fusion}
    \vspace{-10pt}
\end{figure}

In this section, we describe the key components of our Dynamic Virtual-Real Simulation Platform, which integrates virtual-real fusion and dynamic scene generation to support advanced robotic research. These two capabilities are designed to address the limitations of existing simulation platforms, enabling more effective training and evaluation of algorithms in real world conditions. By combining high-fidelity virtual simulation with real-time interactions and dynamic scene modeling, DVS provides a comprehensive environment for testing mobile robots.

\subsection{Virtual-Real Fusion for Seamless Interaction}

Virtual-real fusion is a core feature of DVS, enabling precise bidirectional synchronization between the virtual and physical environments. This synchronization is critical for ensuring that algorithms trained in the virtual world can be directly applied to physical robots, thus reducing the sim-to-real gap.

The virtual-real fusion module consists of two primary components: object pose alignment and robot state synchronization. These components work together to ensure that both the objects and robots in the simulation environment align accurately with their real world counterparts.

\subsubsection{Object Pose Synchronization}

Object pose synchronization is a critical feature for bridging the gap between virtual and real environments, enabling accurate interactions between robots and their surroundings in both domains. In DVS, we achieve precise synchronization using 14 motion capture cameras, which provide real-time tracking with 0.1 mm positional accuracy and 0.1 ° rotational precision. This allows for high-fidelity pose alignment, essential for ensuring that physical objects, such as robot end-effectors, align accurately with their virtual counterparts in simulation.

The synchronization process begins with extrinsic calibration of the motion capture system. By calibrating the system's extrinsic parameters, we can establish a unified world coordinate system that aligns the virtual and real spaces. This calibration is achieved through the following transformation:
\begin{equation}
T_{\text{virtual}} = R \cdot T_{\text{real}} + t
\end{equation}
Where:
\begin{itemize}
    \item $R$ is the rotation matrix derived from the spatial calibration process, defining how the real world orientation maps to the virtual space.
    \item $T_{\text{real}}$ is the translation vector representing the position of the real world object.
    \item $t$ is the translation vector that compensates for any misalignment, ensuring that both spaces share a common origin.
\end{itemize}

Through this method, the physical object trajectories, such as those of a robot's end-effector, are directly mapped into the virtual environment. This enables precise interaction with virtual objects, improving the realism of simulations and ensuring the accuracy of robotic tasks that require interaction between the real and virtual worlds.

\subsection{Dynamic Scene Generation}
The dynamic scene generation module of DVS significantly enhances the realism and complexity of training environments, creating scenarios that more accurately reflect real world conditions.  This module incorporates dynamic pedestrian agents and mobile robotic proxies, both of which are key to simulating the unpredictability and complexity of real world environments.

\subsubsection{Dynamic Pedestrian Plug-in}
DVS features a pedestrian simulation plugin that introduces human-like agents into the virtual environment. These agents feature variable motion accelerations and socially compliant avoidance behaviors, enabling dense, collision-free movement. By incorporating mechanisms such as variable speeds, random spawning points, and obstacle avoidance, the system realistically simulates human interaction dynamics. Adding dynamic pedestrians to different static scenes allows our platform to closely replicate crowded environments like supermarkets and busy restaurants, significantly enhancing the realism of indoor simulations. This provides a richer learning environment for developing mobile robots’ navigation and collaboration capabilities in human-populated spaces.

Such dynamic pedestrian behaviors are crucial for training indoor mobile robots in navigation and human-robot interaction tasks. The human-like agents interact with robots in real time, enabling researchers to collect diverse and realistic datasets for optimizing navigation strategies, path planning, and collaborative algorithms. This feature is particularly valuable for enhancing robots' ability to adapt to sudden environmental changes or unpredictable pedestrian behaviors.

\subsubsection{Multi-Robot Plug-in}
In addition to pedestrian simulation, DVS supports the integration of multiple mobile robotic agents within the same environment. This capability allows researchers to study multi-robot collaboration and competition in dynamic settings. Simulation of multiple robots operating in close proximity enables the development of cooperative algorithms for tasks such as resource sharing, coordinated navigation, and joint manipulation.

The ability to simulate multi-robot environments in dynamic, cluttered spaces is critical for advancing robotics research. By mimicking real world challenges such as managing crowded environments or dealing with unexpected obstacles, DVS helps researchers develop more robust algorithms that can handle complex tasks in unpredictable settings.

Together, dynamic pedestrians and multi-robot integration ensure that DVS provides a training environment that closely mirrors real world operational conditions. These capabilities are essential for developing robots that can navigate complex spaces, collaborate with humans, and adapt to dynamic changes in their environments.

\section{Applications of DVS Platform}
Our platform supports the full workflow, from data generation to real world validation. In the previous chapter, we introduced two core modules of our system. This chapter discusses the construction of a large-scale virtual-real fusion dataset and explores the experimental data generation process, along with its application in task training and testing.

\subsection{Data Perception and Generation}

\begin{figure}[h]
    \centering
    \includegraphics[width=\linewidth]{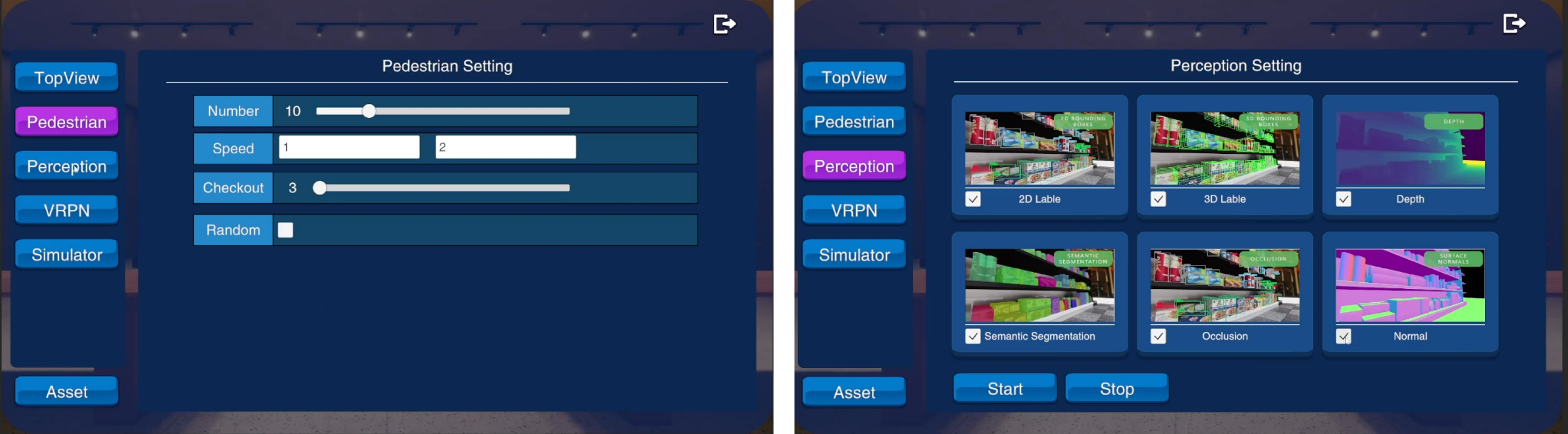}
    \caption{The interactive interface of the simulation platform: The left panel adjusts dynamic pedestrian parameters while the right selects perception data types.}
    \label{fig_ui}
    \vspace{-10pt}
\end{figure}

In robotics research, virtual environments provide clear task representations, enabling agents to perform tasks in controlled settings. Data generation is a core feature of simulation platforms. As shown in Fig.~\ref{fig_ui}, our platform facilitates the generation and processing of various data formats, including RGB images, depth maps, 2D/3D bounding boxes, semantic and instance segmentation, and trajectory data, all via a user-friendly interface \cite{THUD24ICRA}. These data types support foundational tasks and enable complex research scenarios, such as path planning \cite{Navigation24IROS}, relocation\cite{FuseNet24RAL}, and grasping \cite{ZhangHao23ICRA}. 

To enhance data quality and usability in simulation, we optimize the data generation pipeline by ensuring smooth camera trajectories and precise depth-to-RGB alignment. Specifically, we employ Bézier curves to generate smooth camera motion, minimizing abrupt directional changes—particularly at trajectory corners—which significantly improves frame-to-frame feature matching and point cloud reconstruction. Depth data is temporally aligned with RGB frames to guarantee precise synchronization, which is critical for multi-sensor fusion and accurate scene modeling.

We evaluated the impact of these optimizations by collecting camera trajectories both with and without smoothing, and analyzed feature matching performance using LightGlue\cite{lindenberger2023lightglue} and SuperPoint\cite{detone2018superpoint}. The results in Table \ref{tab:future_points} show that smoothed trajectories yield a higher number of feature correspondences between frames, particularly in small and cluttered indoor environments such as bedrooms. As demonstrated in our experiments, these improvements in camera handling directly enhance downstream tasks like 3D reconstruction and semantic segmentation, supporting more robust and scalable robotics research as task complexity increases.

\begin{table}[b]
  \centering
  \small 
  \caption{Camera Trajectories With and Without Smoothing}
  \label{tab:future_points}
  \renewcommand{\arraystretch}{1.2} 
  \setlength{\tabcolsep}{8pt} 
  \begin{tabular}{@{}lcc@{}}
  \toprule
  \textbf{Scene} & \textbf{Trajectory Type} & \textbf{Avg. Features} $\uparrow$ \\
  \midrule
  \multirow{2}{*}{Bedroom} & Straight & 751.35 \\
          & Smooth & 1484.29 \\
  \cmidrule(r){1-3}
  \multirow{2}{*}{Livingroom} & Straight & 1190.47 \\
             & Smooth & 1631.73 \\
  \bottomrule
  \end{tabular}
\end{table}

\subsection{Robotic Tasks Learning}

\begin{figure}[t]
    \centering
    \includegraphics[width=\linewidth]{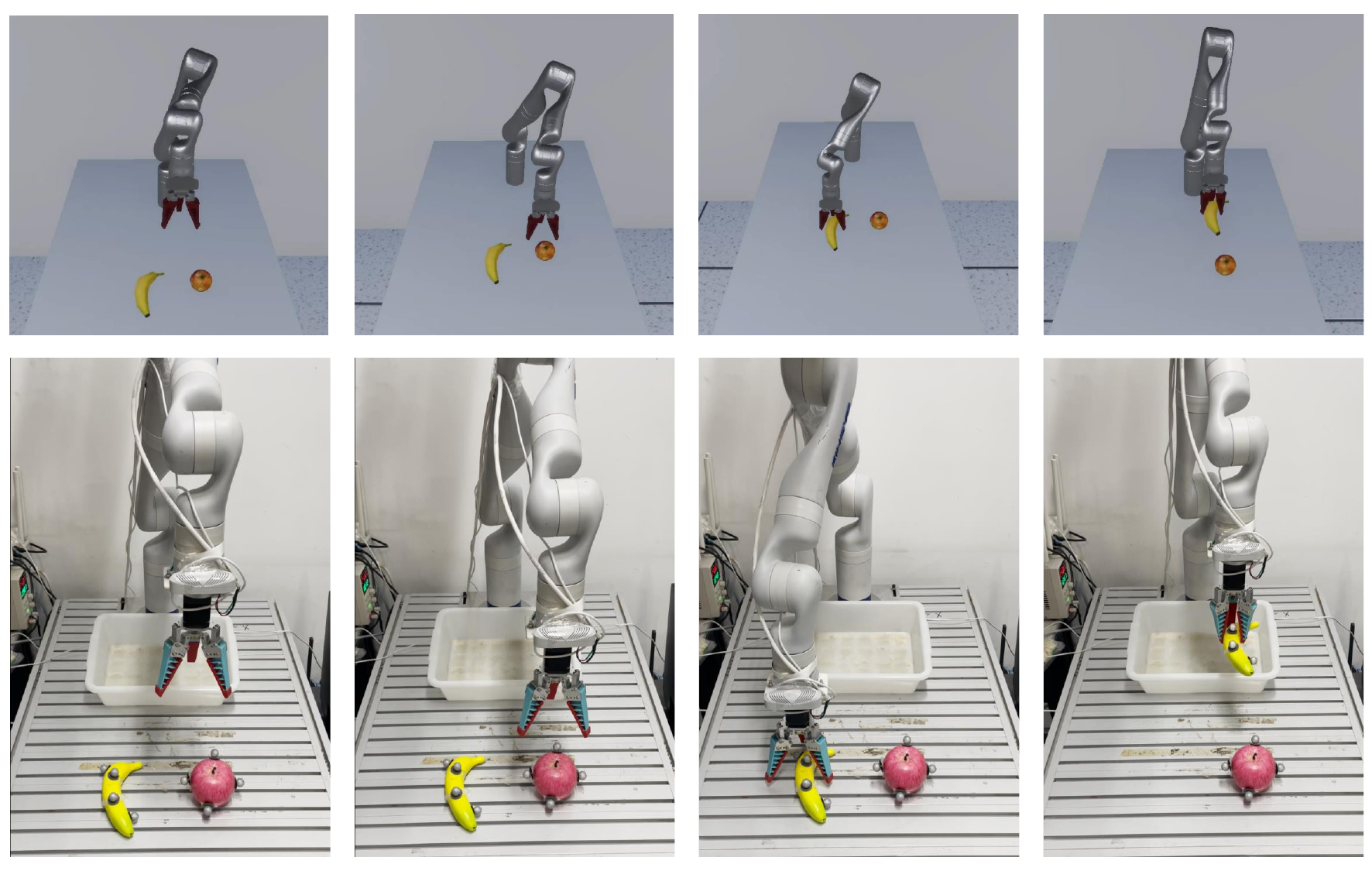}
    \caption{The robotic arm is interrupted while executing Prompt A and is requested to execute Prompt B. The first row shows the robotic arm in the virtual platform, and the second row shows the real robotic arm.}
    \label{fig_grasping}
    \vspace{-15pt}
\end{figure}

\subsubsection{Virtual-Real Intervention Grasping}

\begin{table}[b]
\centering
\caption{Task Success Rates by Module and Prompt Order}
\label{tab:task_success_rates}
\begin{tabular}{lcccrr}
\toprule
\multirow{2}{*}{\makecell{Module}} & \multicolumn{2}{c}{Prompt Order} & \multicolumn{2}{c}{Success Rate (\%)} \\
\cmidrule(lr){2-3} \cmidrule(lr){4-5}
 & First & Second & First Task & Second Task \\
\midrule
\multirow{2}{*}{OpenVLA-7B} 
           & A & B & 0.0 & 100.0 \\
           & B & A & 0.0 & 90.0 \\ \addlinespace[0.3em]
\multirow{2}{*}{RDT-1B}     
           & A & B & 0.0 & 80.0 \\
           & B & A & 0.0 & 90.0 \\
\bottomrule
\end{tabular}
\end{table}

A key weakness of learned policies in robotic manipulation \cite{SDNet24IROS, PPRNet++22TASE, PPR-Net19IROS, ZhangHao23ICRA} is that their success rate in task execution is low when deployed in practice, even with domain adaptation \cite{ZhaoLiang23IROS}. In heterogeneous deployments using only pretrained weights, the success rate of robots performing tasks across different models tends to approach zero. Even when data collection for specific tasks is done using the robot being deployed and fine-tuned, the success rate of task execution is still only around 90$\%$, making it difficult to apply in the industry.
However, due to the characteristics of our platform, which includes virtual-real mapping and benchmark alignment between the virtual environment and the real world, and the fact that the robot has a ROS communication interface, we can supervise and intervene in the robot's tasks in the real world through the platform to improve the success rate of task execution.
We set up experimental conditions based on the common manipulation task of grasping. As shown in Fig.~\ref{fig_grasping}, in order to reflect the characteristics of our platform supervision and intervention, we provide the gripper with wrong instructions at the beginning of the experiment, and interrupt the task and provide new tasks based on virtual scenes through the platform when the gripper is performing the task.
We utilized a seven-degree-of-freedom Kinova Gen3 robotic arm to collect nearly a hundred grasping data points on a planar surface. The data was then fine-tuned on the pre-trained models released by OpenVLA-7B\cite{kim2024openvla} and RDT-1B\cite{liu2024rdt}, enabling our robotic arm to achieve a high success rate in performing tasks in specific scenarios. At the beginning of the experiment, we provided the robotic arm with prompts to grasp an apple and a banana, and midway through task execution, we interrupted the task on the platform and assigned a new task.The experiment demonstrated that our platform effectively intervened in the robotic arm’s task execution. The experimental results are shown in the Table \ref{tab:task_success_rates}.
Prompts: A: "Pick up the apple"; B: "Pick up the banana."

\subsubsection{Real to Sim to Real Learning}
In autonomous robotic task execution without human intervention, integrating physical robots and their sensory systems into a virtual-real fusion architecture serves as a key technical strategy for improving task success rates. The real-world component of our platform includes the physical robot and sensing devices used to collect feedback. A notable application of virtual-real synchronization is digital twin monitoring, which enables continuous evaluation and refinement of algorithms using real-world data—effectively bridging the sim-to-real gap.

To achieve precise synchronization, we employ ROS2 to align the real robot’s motion, controlled via MoveIt, with that of its virtual counterpart. This ensures the virtual environment reflects physical priors such as joint friction and mechanical latency. We validated this approach in a Virtual-Real Assisted (VLA) grasping task by comparing standard sim-to-real transfer with a fine-tuned version incorporating real-world feedback. As shown in Table~\ref{tab:Finetuning}, the virtual-real fusion method led to significantly better performance, highlighting the value of real-time feedback in simulation refinement.

\begin{table}[t]
  \centering
  \caption{Comparative with Different Finetuning Data }
  \label{tab:Finetuning}
  \setlength{\tabcolsep}{6pt}
  \footnotesize
  \begin{tabular}{@{}l c l c@{}}
  \toprule
  \textbf{Task} & \textbf{Trials} & \textbf{Finetune Data} & \textbf{Successes} \\
  \midrule
  \multirow{2}{*}{\smash{\begin{tabular}{@{}>{\raggedright}p{3.3cm}@{}}Pick the apple and place \\ at the target point\end{tabular}}} 
  & 10 & virtual & 6 \\
  & 10 & virtual-real & 9 \\[3pt]
  
  \multirow{2}{*}{\smash{\begin{tabular}{@{}>{\raggedright}p{3.3cm}@{}}Pick the banana and place \\ at the target point\end{tabular}}}
  & 10 & virtual & 4 \\
  & 10 & virtual-real & 8 \\
  \bottomrule
  \end{tabular}
\end{table}

To support accurate virtual-real synchronization, we use a motion capture system for high-precision calibration between real and virtual coordinate systems. This calibration is essential for aligning the physical robot’s position with its virtual counterpart, correcting errors such as odometry drift and IMU noise. For tasks that are less sensitive to positional accuracy, we leverage the built-in mapping capabilities of the Quest system to achieve approximate alignment—a low-cost yet sufficiently accurate alternative for many practical applications.

\subsubsection{Dynamic Indoor Pedestrian Trajectory Prediction}

\begin{figure}[b]
    \centering
    \includegraphics[width=\linewidth]{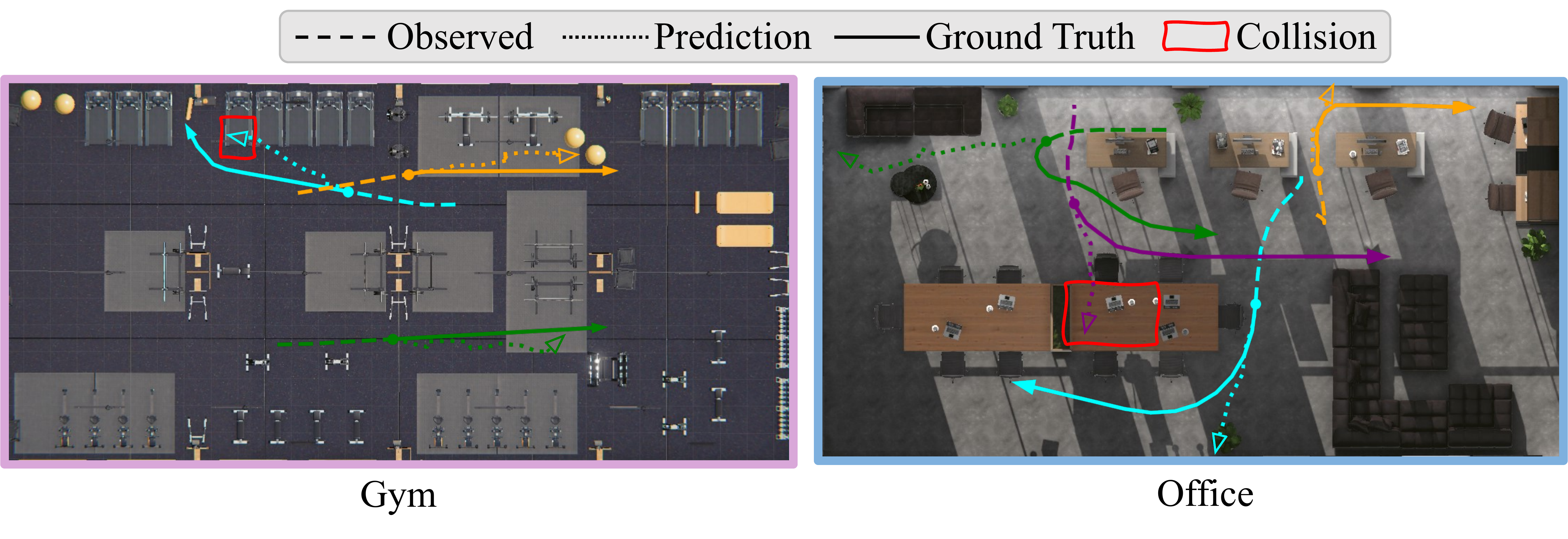}
    \caption{Visualization of pedestrian trajectory prediction, where each color represents a different pedestrian. The accuracy of the prediction is higher when the predicted trajectory (short dashed line) closely aligns with the ground truth (GT, solid line). In environments with dense static obstacles, such as indoors, the predicted future trajectory may result in collisions (red rectangular box).}
    \label{fig_trajectory}
    \vspace{-0pt}
\end{figure}

Pedestrian trajectory prediction aims to forecast future trajectories based on observed trajectories, while considering complex interactions and environmental layouts \cite{Trajectory24RAL}. It serves as a crucial connection between the perception system and the planning system.

Three trajectory prediction algorithms, i.e. STGAT \cite{huang2019stgat}, Trajectron++ \cite{salzmann2020trajectron++} and TUTR \cite{shi2023trajectory}, are tested on our synthetic indoor scenes (Gym, Office and Supermarket) as well as the official public outdoor dataset (ETH \cite{eth2009}). We use ADE (Average Displacement Error) and FDE (Final Displacement Error) as evaluation metrics, where lower ADE and FDE values indicate better performance. The experimental results are depicted in Table \ref{tab: ped traj predict}. Additionally, to analyze pedestrian movement patterns and collision avoidance strategies, we selected two dense indoor scenes (Gym and Office) and visualized the predicted trajectories in Fig.~\ref{fig_trajectory}.

Overall, all three methods experience a significant performance decrease when applied to indoor scenes compared to the outdoor ETH scene. Specifically, the ADE for STGAT decreases from 0.79 to 1.42 (79.7$\%$) when generalizing from the ETH scene to the Supermarket scene, while the FDE for STGAT decreases from 1.48 to 2.88 (94.5$\%$) in the same scenario.  We analyze this performance drop from three perspectives. First, compared to outdoor scenes, narrow indoor spaces are often filled with numerous static obstacles, which can interfere with human trajectory decision-making and lead to collisions. Second, indoor human interactions are more frequent due to communication or obstacles caused by people standing in the way, making predictions more challenging. Third, indoor spaces are generally smaller than outdoor environments, with pedestrian trajectories being less spread out, making predictions more sensitive to small positional changes. If the model was trained in larger, more open outdoor spaces, it may not have learned to adapt to the smaller, more dynamic movements of indoor environments.

The results also underscore the importance of robust spatial-temporal modeling in trajectory prediction tasks. The transformer-based architecture of TUTR appears particularly well suited to capture intricate interactions over time, which leads to its superior performance. Trajectoron provides a balance of stability and accuracy, but lags behind in highly dynamic environments. In contrast, STGAT’s graph-based approach, while effective in simpler scenarios, struggles in complex environments, highlighting its limitations in handling high-dimensional spatial-temporal variability. These findings offer valuable insights for future research, emphasizing the need for models that can generalize effectively across diverse scenarios while maintaining low computational overhead.

\begin{table}[!t]
\centering
\caption{Experiments on Pedestrian Trajectory Prediction. Gym, Office and Supermarket are our synthetic indoor scenes, while ETH \cite{eth2009} is the official public outdoor dataset.}
\label{tab: ped traj predict}
\begin{tabular}{cccc}
    \toprule
    \textbf{Scene} & \textbf{Method} & \textbf{ADE} $\downarrow$ & \textbf{FDE} $\downarrow$ \\ \midrule
    \multirow{3}{*}{Gym} & STGAT & 1.39 & 3.01 \\
     & Trajectron++ & 0.59 & 1.02 \\
     & TUTR & 0.70 & 1.19 \\ \midrule
    \multirow{3}{*}{Office} & STGAT & 1.38 & 2.75 \\
     & Trajectron++ & 0.89 & 1.60 \\
     & TUTR & 0.81 & 1.40 \\ \midrule
    \multirow{3}{*}{Supermarket} & STGAT & 1.42 & 2.88 \\
     & Trajectron++ & 0.96 & 1.82 \\
     & TUTR & 0.83 & 1.50 \\ \midrule
    \multirow{3}{*}{ETH} & STGAT & 0.79 & 1.48\\
     & Trajectron++ & 0.52 & 0.97 \\
     & TUTR & 0.43 & 0.83 \\ \bottomrule
    \vspace{-10pt}
\end{tabular}
\end{table}

\begin{table*}[t]
    \centering
    \caption{Experiments on Social Navigation}
    \renewcommand{\arraystretch}{1.5} 
    \label{tab:crowdnav-results}
    \resizebox{0.8\textwidth}{!}{
        \begin{tabular}{@{}ll cc cc cc @{}}
        \toprule
        \multirow{2}{*}{\textbf{Scene}} & \multirow{2}{*}{\textbf{Metric}} & \multicolumn{3}{c}{\textbf{HumanNumber=10 / 15 / 20}}\\
        \cmidrule(lr){3-5} 
         & & \textbf{ORCA} & \textbf{DS-RNN} & \textbf{AttnGraph} \\
        \midrule
        \multirow{3}{*}{Restaurant} 
        & SuccessRate $\uparrow$ & 0.78 / 0.74 / 0.62 & 0.82 / 0.76 / 0.68 & 0.83 / 0.77 / 0.67 \\
        & CollisionRate $\downarrow$ & 0.01 / 0.06 / 0.08 & 0.01 / 0.05 / 0.06 & 0.02 / 0.06 / 0.07 \\
        & NavigationTime $\downarrow$ & 42.50 / 43.59 / 43.90 & 37.48 / 44.96 / 45.24 & 39.68 / 44.81 / 49.65 \\
        \midrule
        
        \multirow{3}{*}{Store}
        & SuccessRate $\uparrow$ & 0.96 / 0.85 / 0.51 & 0.98 / 0.81 / 0.75 & 0.98 / 0.87 / 0.79 \\
        & CollisionRate $\downarrow$ & 0.03 / 0.04 / 0.06 & 0.01 / 0.07 / 0.09 & 0.02 / 0.04 / 0.06 \\
        & NavigationTime $\downarrow$ & 40.39 / 47.62 / 46.47 & 34.21 / 39.51 / 39.97 & 41.56 / 43.75 / 48.22 \\
        \bottomrule
        \end{tabular}
    }
\end{table*}

\subsubsection{Dynamic Indoor Social Navigation}
Social navigation refers to the process by which agents use social cues and prior experiences to determine paths, make decisions, and navigate complex environments. It incorporates interpersonal and collective information, including behavioral patterns, real-time human feedback, and established social norms.

We evaluated three indoor social navigation algorithms ORCA \cite{van2011reciprocal}, DS-RNN \cite{liu2021decentralized}, and AttnGraph \cite{liu2023intention} within synthetic indoor restaurant and store environments in our platform with varying levels of dynamic complexity. Performance was quantified using three primary metrics: Success Rate, average Navigation Time (in seconds) for successful episodes, and Collision Rate with other humans. The experimental results are shown in Table \ref{tab:crowdnav-results}.

The results indicate that with increasing dynamic scene complexity, the performance of all three algorithms deteriorates to varying extents, consistent with our expectations. Specifically, as the number of dynamic pedestrians increases, the success rate decreases, whereas the collision rate and navigation time increase, each to different extents. This demonstrates the influence of dynamic scene complexity on algorithm performance. This further highlights that our online simulation platform is well suited for closed-loop training and testing.

\section{Conclusion}
We propose a dynamic virtual-real simulation platform that integrates configurable pedestrian behavior simulation, large-scale indoor environments, optical motion capture, and ROS-based bidirectional virtual-reality communication. The platform introduces two major innovative modules for virtual reality integration, overcoming current limitations in robotic simulation systems for dynamic scenarios and real world deployment. Experimental results show that DVS supports navigation and human-robot interaction research, achieving closed-loop performance in real world missions. Future work will focus on integrating haptic feedback, developing AI-driven intervention strategies, and improving compatibility with industrial robotic arms. This platform creates a new paradigm for closed-loop virtual-reality interaction, advancing human-robot collaboration and dynamic environment adaptation.

\section*{Acknowledgments}
This work was supported by the Shenzhen Science and Technology Major Project (KJZD20230923115503007), Shenzhen Major Undertaking Plan (CJGJZD20240729141702003), and Joint Research Center of Tsinghua-Pudu Intelligent Service Robotic Technology.


\bibliographystyle{plainnat}
\bibliography{references}

\end{document}